\documentclass{article}

\usepackage{microtype}
\usepackage{graphicx}
\usepackage{subfigure}
\usepackage{booktabs} 
\usepackage{array, multirow}
\usepackage{amsfonts}       
\usepackage{hyperref}
\usepackage{amsmath}

\newcommand\norm[1]{\left\lVert#1\right\rVert}

\usepackage[accepted]{icml2019}

\icmltitlerunning{Modeling of Random Noise during Training improves Adversarial Robustness}

\begin{document}

\twocolumn[
\icmltitle{Implicit Generative Modeling of Random Noise during Training improves Adversarial Robustness}




\begin{icmlauthorlist}
\icmlauthor{Priyadarshini Panda}{to}
\icmlauthor{Kaushik Roy}{to}

\end{icmlauthorlist}

\icmlaffiliation{to}{Department of Electrical \& Computer Engineering, Purdue University, West Lafayette, USA.\\ A preliminary version of this work is accepted at ICML 2019 Workshop on Uncertainty and Robustness in Deep Learning}

\icmlcorrespondingauthor{P. Panda}{pandap@purdue.edu}

\icmlkeywords{Machine Learning, ICML}

\vskip 0.3in
]



\printAffiliationsAndNotice{}  

\begin{abstract}
We introduce a \underline{No}ise-based prior \underline{L}earning (NoL) approach for training neural networks that are intrinsically robust to adversarial attacks. We find that the implicit generative modeling of random noise with the same loss function used during posterior maximization, improves a model's understanding of the data manifold furthering adversarial robustness. We evaluate our approach's efficacy and provide a simplistic visualization tool for understanding adversarial data, using Principal Component Analysis. Our analysis reveals that adversarial robustness, in general, manifests in models with higher variance along the high-ranked principal components. We show that models learnt with our approach perform remarkably well against a wide-range of attacks. Furthermore, combining NoL with state-of-the-art adversarial training extends the robustness of a model, even beyond what it is adversarially trained for, in both white-box and black-box attacks. \textit{Source Code available at \href{https://github.com/panda1230/Adversarial_NoiseLearning_NoL.git}{panda1230/Adversarial\_NoiseLearning\_NoL}}.
\end{abstract}

\section{Introduction}
Despite surpassing human performance on several perception tasks, Machine Learning (ML) models remain vulnerable to \textit{adversarial examples}: slightly perturbed inputs that are specifically designed to fool a model during test time \citep{biggio2013evasion,szegedy2013intriguing, goodfellow2014explaining, papernot2016limitations}. Recent works have demonstrated the security danger adversarial attacks pose across several platforms with ML backend such as computer vision \citep{szegedy2013intriguing, goodfellow2014explaining, moosavi2016deepfool, kurakin2016adversarial, liu2016delving}, malware detectors \citep{laskov2014practical, xu2016automatically, grosse2016adversarial, hu2017generating} and gaming environments \citep{huang2017adversarial, behzadan2017vulnerability}. Even worse, adversarial inputs \textit{transfer} across models: same inputs are misclassified by different models trained for the same task, thus enabling simple \textit{Black-Box} (BB) \footnote{\label{note1}BB (WB): attacker has no (full) knowledge of the target model parameters}attacks against deployed ML systems \citep{papernot2017practical}.

Several works \citep{krotov2017dense, papernot2016distillation, cisse2017parseval} demonstrating improved adversarial robustness have been shown to fail against stronger attacks \citep{athalye2018obfuscated}.  
The state-of-the-art approach for BB defense is ensemble adversarial training that augments the training dataset of the target model with adversarial examples transferred from other pre-trained models \citep{tramer2017ensemble}. \cite{madry2017towards} showed that models can even be made robust to \textit{White-Box} (WB)\textsuperscript{\ref{note1}} attacks by closely maximizing the model's loss with Projected Gradient Descent (PGD) based adversarial training. Despite this progress, errors still appear for perturbations beyond what the model is adversarially trained for \citep{sharma2017breaking}. 

There have been several hypotheses explaining the susceptibility of ML models to such attacks. The most common one suggests that the overly linear behavior of deep neural models in a high dimensional input space causes adversarial examples \citep{goodfellow2014explaining, lou2016foveation}. Another hypothesis suggests that adversarial examples are off the data manifold \citep{goodfellow2016deep, song2017pixeldefend, lee2017generative}. Combining the two, we infer that excessive linearity causes models to extrapolate their behavior beyond the data manifold yielding pathological results for slightly perturbed inputs. A question worth asking here is: \textit{Can we improve the viability of the model to generalize better on such out-of-sample data?} 

\begin{figure*}[t!]
\centering
\includegraphics[width=\linewidth]{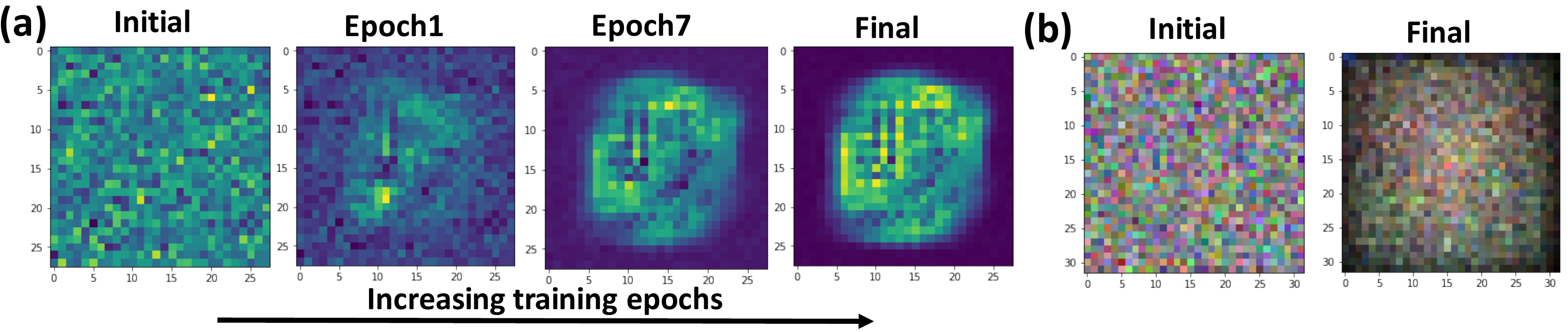}
\caption{(a) Noise learnt with NoL on MNIST data- (b) Noise learnt with NoL on CIFAR10 data- with mini-batch size =64.The template shown is the mean across all 64 noise templates.}
\label{fig1}
\end{figure*}

In this paper, we propose \textit{Noise-based Prior Learning (NoL)}, wherein we introduce multiplicative noise into the training inputs and optimize it with Stochastic Gradient Descent (SGD) while minimizing the overall cost function over the training data. Essentially, the input noise (randomly initialized at the beginning) is gradually learnt during the training procedure. As a result, the noise approximately models the input distribution to effectively maximize the likelihood of the class labels given the inputs. Fig. \ref{fig1} (a) shows the input noise learnt during different stages of training by a simple convolutional network ($ConvNet2$ architecture discussed in Section 3 below), learning handwritten digits from MNIST dataset \citep{lecun1998gradient}. We observe that the noise gradually transforms and finally assumes a shape that highlights the most dominant features in the MNIST training data. For instance, the MNIST images are centered digits on a black background. Noise, in fact, learnt this centered characteristic. Fig. \ref{fig1} suggests that noise discovers some knowledge about the input/output distribution during training. Fig. \ref{fig1} (b) shows the noise learnt with NoL on colored CIFAR10 images \citep{krizhevsky2009learning} (on ResNet18 architecture \citep{he2016deep}), which reveals that noise template (also RGB) learns prominent color blobs on a greyish-black background, that de-emphasizes background pixels.

A recent theory \citep{gilmer2018adversarial} suggests that adversarial examples (off manifold misclassified points) occur in close proximity to randomly chosen inputs on the data manifold that are, in fact, correctly classified. With NoL, we hypothesize that the model learns to look in the vicinity of the on-manifold data points and thereby incorporate more out-of-sample data (without using any direct data augmentation) that, in turn, improves its’ generalization capability in the off-manifold input space. We empirically evaluate this hypothesis by visualizing and studying the relationship between the adversarial and the clean inputs using Principal Component Analysis (PCA). Examining the intermediate layer's output, we discover that models exhibiting adversarial robustness yield significantly lower distance between adversarial and clean inputs in the Principal Component (PC) subspace.
We further harness this result to establish that NoL noise modeling, indeed, acquires an improved realization of the input/output distribution characteristics that enables it to generalize better. 
To  further substantiate our hypothesis, we also show that NoL globally reduces the dimensionality of the space of adversarial examples \citep{tramer2017space}. We evaluate our approach on classification tasks such as MNIST, CIFAR10 and CIFAR100 and show that models trained with NoL are extensively more adversarially robust. We also show that combining NoL with ensemble/PGD adversarial training significantly extends the robustness of a model, even beyond what it is adversarially trained for, in both BB/WB attack scenarios. 

\section{Noise-based Prior Learning}
\subsection{Approach}
The basic idea of NoL is to inject random noise with the training data, continually minimizing the overall loss function by learning the parameters, as well as the noise at every step of training. The noise, $N$, dimensionality is same as the input, $X$, that is, for a $32\times32\times 3$ sized image, the noise is $32\times 32\times 3$. In all our experiments, we use mini-batch SGD optimization. Let's assume the size of the training minibatch is $m$ and the number of images in the minibatch is $k$, then, total training images are $m\times k$. Now, the total number of noisy templates are equal to the total number of inputs in each minibatch, $k$. Since, we want to learn the noise, we use the same $k$ noise templates across all mini-batches ${1, 2,...,m}$. This ensures that the noise templates inherit characteristics from the entire training dataset. Algorithm 1 shows the training procedure. It is evident from Algorithm 1 that noise learning at every training step follows the overall loss ($\mathcal{L}$, say cross-entropy) minimization that in turn enforces the maximum likelihood of the posterior. 
\begin{algorithm}
\caption{Noise-based Prior Learning of a model $f$ with parameters $\theta$, Loss Function $\mathcal{L}$.}
\textbf{Input:} Input image $X$, Target label $Y$, Noise $N$, Learning rates $\eta, \eta_{noise}$\\
\textbf{Output:} Learnt noise $N$ and parameters $\theta$. \\
 Randomly initialize the parameters $\theta$ and Noise $N: \{N^1,...N^k\}$. \\
 \textbf{repeat}\\
\textbf{for} each minibatch $\{X^{[1]},...,X^{[m]}\}$ \\
\hspace*{1em} Input $X$= $\{X^1, ..., X^k\}$\\
 \hspace*{1em} New input $X’$= $\{X^1 \times N^1, ..., X^k \times N^k\}$\\
\hspace*{1em} \textbf{Forward Propagation:} $\hat{Y}$ = $f(X’;\theta)$\\
\hspace*{1em} \textbf{Compute loss function:} $\mathcal{L}(\hat{Y}, Y)$\\
\hspace*{1em} \textbf{Backward Propagation:}\\
\hspace*{1em} \hspace*{2mm} $\theta$ = $\theta$ - $\eta \nabla_{\theta}\mathcal{L}$; $N$ = $N$ - $\eta_{noise} \nabla_{N}\mathcal{L}$ \\
\textbf{end}\\
 \textbf{until} training converges

\end{algorithm}

\begin{figure*}[h]
\centering
\includegraphics[width=\textwidth]{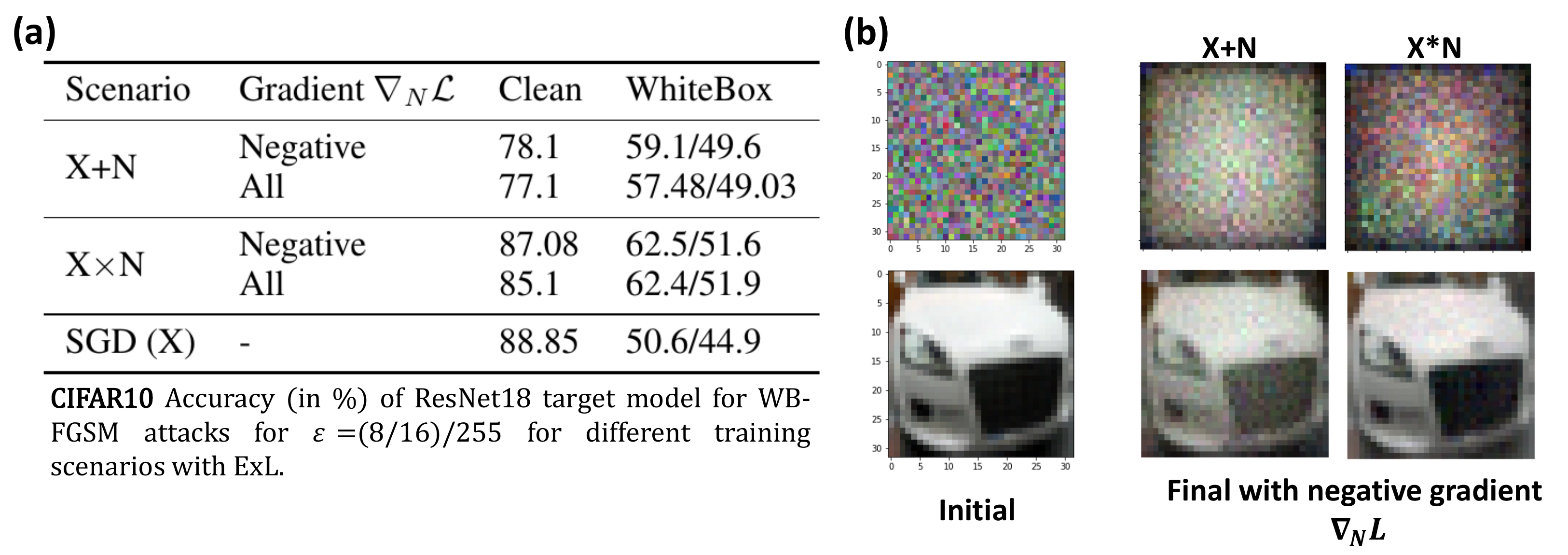}
\caption{For multiplicative and additive noise training scenarios- (a) -accuracy comparison of NoL with SGD (b) -RGB noise template learnt with NoL on CIFAR10 data. In (b), a sample training image of a `car' before and after training with noise is shown. Note, we used the same hyperparameters (batch-size =64, $\eta, \eta_{noise}$ etc.) and same inital noise template across all scenarios during training. Noise shown is the mean across 64 templates.\textsuperscript{\ref{note2}}}
\label{fig2}
\end{figure*}

Since adversarial attacks are created by adding perturbation to the clean input images, we were initially inclined toward using additive noise ($X+N$) instead of multiplicative noise ($X\times N$) to perform NoL. However, we found that NoL training with $X\times N$ tends to learn improved noise characteristics by the end of training. Fig. \ref{fig2} (a) shows the performance results for different NoL training scenarios. While NoL with $X+N$ suffers a drastic $\sim10\%$ accuracy loss with respect to standard $SGD$ on clean data, $X\times N$ yields comparable accuracy. Furthermore, we observe that using only negative gradients for training the noise (i.e. $\nabla_{N}\mathcal{L} \le 0$) during backpropagation with NoL yields best accuracy (and closer to that of standard SGD trained model). Visualizing a sample image with learnt noise after training, in Fig. \ref{fig2} (b), shows $X+N$ disturbs the original image severely, while $X\times N$ has a faint effect, corroborating the accuracy results. Since noise is modeled while conducting discriminative training, the multiplicative/additive nature of noise influences the overall optimization. Thus, we observe that noise templates learnt with $X\times N$ and $X+N$ are very different. We also analyzed the adversarial robustness of the models when subjected to WB attacks created using the Fast Gradient Sign Method (FGSM) for different perturbation levels ($\epsilon$) (Fig. \ref{fig2} (a)). NoL, for both $X\times N$/$X+N$ scenarios, yields improved accuracy than standard SGD. This establishes the effectiveness of the noise modeling technique during discriminative training towards improving a model's intrinsic adversarial resistance. Still, $X\times N$ yields slightly better resistance than $X+ N$. Based upon these empirical studies, we chose to conform to multiplicative noise training in this paper. \footnote{\label{note2}Additional studies on other datasets comparing $X+N$ vs. $X \times N$ with different gradient update conditions can be found in Appendix A. See, experimental details and model description for Fig. \ref{fig2} in Appendix C.1.} Note, WB attacks, in case of NoL, are crafted using the model's parameters as well as the learnt noise $N$.  

In all our experiments, we initialize the noise $N$ from a random uniform distribution in the range $[0.8, 1]$. We select a high range in the beginning of training to limit the corruption induced on the training data due to the additional noise. During evaluation/testing, we take the mean of the learnt noise across all the templates ($(\sum_{i=1}^{k}N_i)/k$), multiply the averaged noise with each test image and feed it to the network to obtain the final prediction.  Next, we present a general optimization perspective considering the maximum likelihood criterion for a classification task to explain adversarial robustness. It is worth mentioning that while Algorithm 1 describes the backpropagation step simply by using gradient updates, we can use other techniques like regularization, momentum etc. for improved optimization. 

It is noteworthy to reiterate that in our NoL approach, we introduce a random set of noise templates (say, $K$ noise templates) to the training data (say, $M$ number of total training instances in the dataset), where, $K << M$. Here, as we learn the noise templates ($N: \{N^1,...N^K\}$) over the duration of training, the noise templates (randomly initialized at the beginning of training) begin to model the input distribution (as seen from Fig. \ref{fig1}). The fact that the same set of noise templates are introduced batch-wise (when training with mini-batch gradient descent) with the training data, enables the prior modeling property. Thus, in Fig. \ref{fig1}, we see the evolution of the noise templates from random distribution in the initial epoch to some shape (characteristic of input data) in the final epoch (as they are learnt with backpropagation $N := N - \nabla_{N}\mathcal{L}$). $NoL$ is in contrast to previous works \cite{noh2017regularizing, fawzi2016robustness, HintonLec}, wherein, random noise is injected to the training dataset, which can be viewed as having same number of noise templates as the number of training instances, that is, $K = M$. Here, the noise templates are simply used to perturb the training data to impose some regularization effect during training. These noise templates are `not learnt' unlike $NoL$. In such cases, each batch (when training with mini-batch gradient descent) of training data sees a different set of random noise templates drawn from a given distribution. Referring to Fig \ref{fig1}, with random noise injection as \cite{noh2017regularizing, fawzi2016robustness, HintonLec}, we will see that the noise template remains random as shown in the initial epoch throughout the training process till the final epoch. 

\subsection{Adversarial Robustness from Likelihood Perspective}
Given a data distribution $D$ with inputs $X\in \mathbb{R}^d$ and corresponding labels $Y$, a classification/discriminative algorithm models the conditional distribution $p(Y|X;\theta)$ by learning the parameters $\theta$. Since $X$ inherits only the on-manifold data points, a standard model thereby becomes susceptible to adversarial attacks. For adversarial robustness, inclusion of the off-manifold data points while modeling the conditional probability is imperative. An adversarially robust model should, thus, model $p(Y|X, \mathbb{A};\theta)$, where $\mathbb{A}$ represents the adversarial inputs. Using Bayes rule, we can derive the prediction obtained from posterior modeling from a generative standpoint as:
\begin{align}
\underset{Y}{argmax}\hspace{0.5mm}p(Y|X,\mathbb{A}) & = \underset{Y}{argmax}\hspace{0.5mm}\frac{p(\mathbb{A}|X,Y) p(X,Y)}{p(X, \mathbb{A})} \\
& = \underset{Y}{argmax}\hspace{0.5mm}p(\mathbb{A}|X,Y) p(X|Y) p(Y)
\label{eq1}
\vspace{-1mm}
\end{align}

The methods employing adversarial training \citep{tramer2017ensemble,kurakin2016adversarial,madry2017towards} directly follow the left-hand side of Eqn. \ref{eq1} wherein the training data is augmented with adversarial samples ($A\in \mathbb{A}$). Such methods showcase adversarial robustness against a particular form of adversary (e.g. $\ell_{\infty}$-norm bounded) and hence remain vulnerable to stronger attack scenarios. In an ideal case, $A$ must encompass all set of adversarial examples (or the entire space of off-manifold data) for a concrete guarantee of robustness. However, it is infeasible to anticipate all forms of adversarial attacks during training. From a generative viewpoint (right-hand side of Eqn. \ref{eq1}), adversarial robustness requires modeling of the adversarial distribution while realizing the joint input/output distribution characteristics ($p(X|Y), p(Y)$). Yet, it remains a difficult engineering challenge to create rich generative models that can capture these distributions accurately. Some recent works leveraging a generative model for robustness use a PixelCNN model \citep{song2017pixeldefend} to detect adversarial examples, or use Generative Adversarial Networks (GANs) to generate adversarial examples \citep{samangouei2018defense}. But, one might come across practical difficulties while implementing such methods due to the inherent training difficulty. 

With Noise-based Prior Learning, we partially address the above difficulty by modeling the noise based on the prediction loss of the posterior distribution. First, let us assume that the noise ($\mathbb{N}$) introduced with NoL spans a subspace of potential adversarial examples ($\mathbb{N} \subseteq \mathbb{A}$). Based on Eqn. \ref{eq1} the posterior optimization criterion with noise ($\mathbb{N}$) becomes ${argmax_Y} \hspace{0.5mm} p(Y|X,\mathbb{N}) = {argmax_Y}\hspace{0.5mm} p(\mathbb{N}|X,Y) p(X|Y) p(Y)$. The noise learning in NoL (Algorithm 1) indicates an implicit generative modeling behavior, that is constrained towards maximizing $p(\mathbb{N}|X,Y)$ while increasing the likelihood of the posterior $p(Y|X,\mathbb{N})$. We believe that this partial and implicit generative modeling perspective with posterior maximization, during training, imparts an NoL model more knowledge about the data manifold, rendering it less susceptible toward adversarial attacks. 

Intuitively, we can justify this robustness in two ways: First, by integrating noise during training, we allow a model to explore multiple directions within the vicinity of the data point (thereby incorporating more off-manifold data) and hence inculcate that knowledge in its underlying behavior. Second, we note that noise learnt with NoL inherits the input data characteristics (i.e. $\mathbb{N}\subset X$) and that the noise-modeling direction ($\nabla_N \mathcal{L}$) is aligned with the loss gradient, $\nabla_X \mathcal{L}$ (that is also used to calculate the adversarial inputs, $X_{adv} = X + \epsilon sign(\nabla_X \mathcal{L})$). This ensures that the exploration direction coincides with certain adversarial directions improving the model's generalization capability in such spaces. Next, we empirically demonstrate using PCA that, noise modeling indeed embraces some off-manifold data points. Note, for fully guaranteed adversarial robustness as per Eqn. \ref{eq1}, the joint input/output distribution ($p(X|Y), p(Y)$) has to be realized in addition to the noise modeling and $\mathbb{N}$ should span the entire space of adversarial/off-manifold data. We would like to clarify that the above likelihood perspective is our intuition that noise might be inculcating some sort of generaive behavior in the discriminative classification. However, more rigorous theoretical analysis is required to understand the underlying behavior of noise.

\subsection{PC Subspace Analysis for Variance \& Visualization}
\begin{figure*}[h]
\centering
\includegraphics[width=\textwidth]{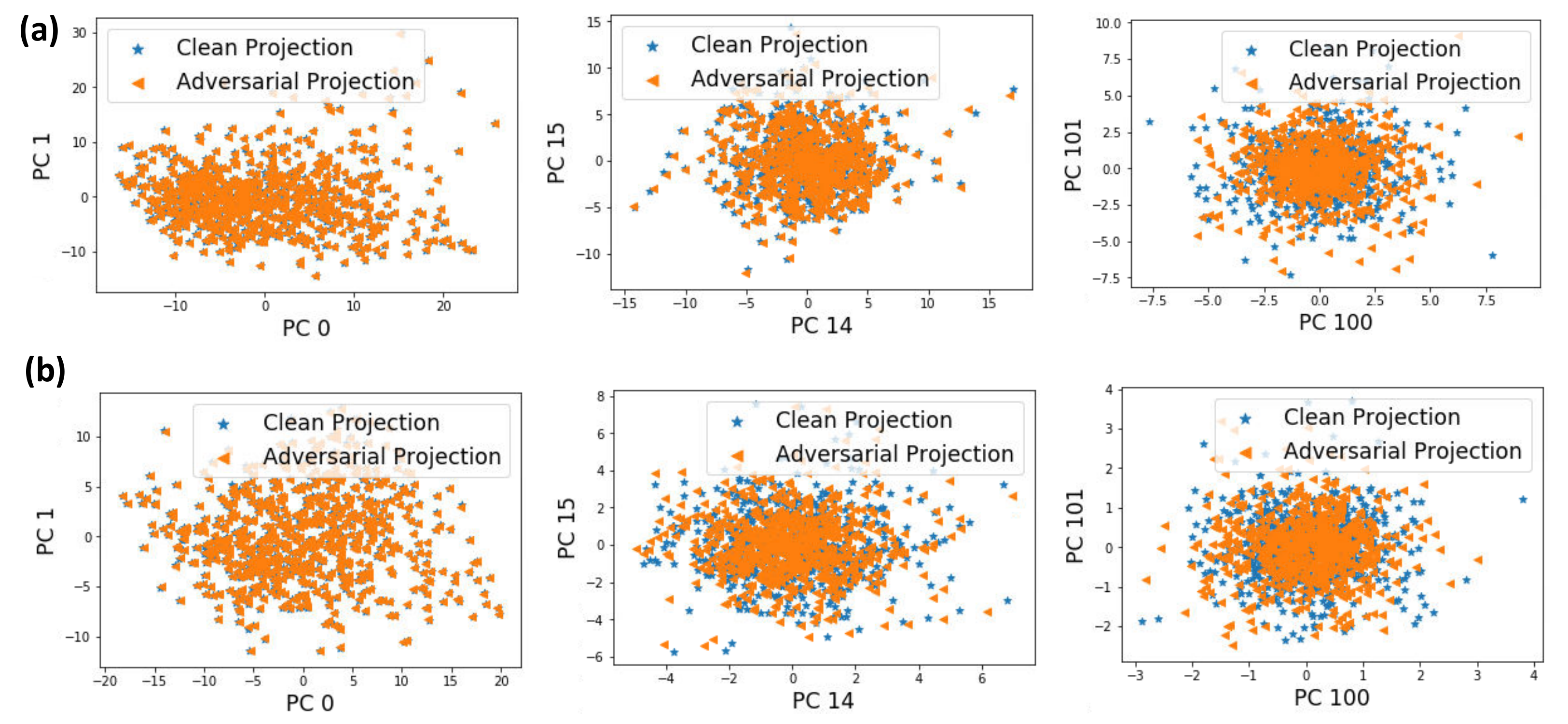}
\caption{Relationship between the model's understanding of adversarial and clean inputs in PC subspace when trained with (a) NoL (b) SGD.}
\label{fig3_1}
\end{figure*}

PCA serves as a method to reduce a complex dataset to lower dimensions to reveal sometimes hidden, simplified structure that often underlie it. Since the learned representations of a deep learning model lie in a high dimensional geometry of the data manifold, we opted to reduce the dimensionality of the feature space and visualize the relationship between the adversarial and clean inputs in this reduced PC subspace. 
Essentially, we find the principal components (or eigen-vectors) of the activations of an intermediate layer of a trained model and project the learnt features onto the PC space. To do this, we center the learned features about zero ($\mathcal{F}$), factorize $\mathcal{F}$ using Singular Value Decomposition (SVD), i.e. $\mathcal{F}=USV^T$ and then transform the feature samples $\mathcal{F}$ onto the new subspace by computing $\mathcal{F}V=US \equiv \mathcal{F}^{PC}$. In Fig. \ref{fig3_1} (b), we visualize the learnt representations of the \textit{Conv1 layer} of a ResNet18 model trained on CIFAR-10 (with standard SGD) along different 2D-projections of the PC subspace in response to adversarial/clean input images. Interestingly, we see that the model's perception of both the adversarial and clean inputs along high-rank PCs (say, PC1- PC10 that account for maximum variance in the data) is alike. As we move toward lower-rank dimensions, the adversarial and clean image representations dissociate. This implies that adversarial images place strong emphasis on PCs that account for little variance in the data. While we note a similar trend with NoL (Fig. \ref{fig3_1} (a)), the dissociation occurs at latter PC dimensions compared to Fig. \ref{fig3_1} (b). A noteworthy observation here is that, adversarial examples lie in close vicinity of the clean inputs for both NoL/SGD scenarios ascertaining former theories of \citep{gilmer2018adversarial}.

\begin{figure*}[h]
\centering
\includegraphics[width=0.8\textwidth]{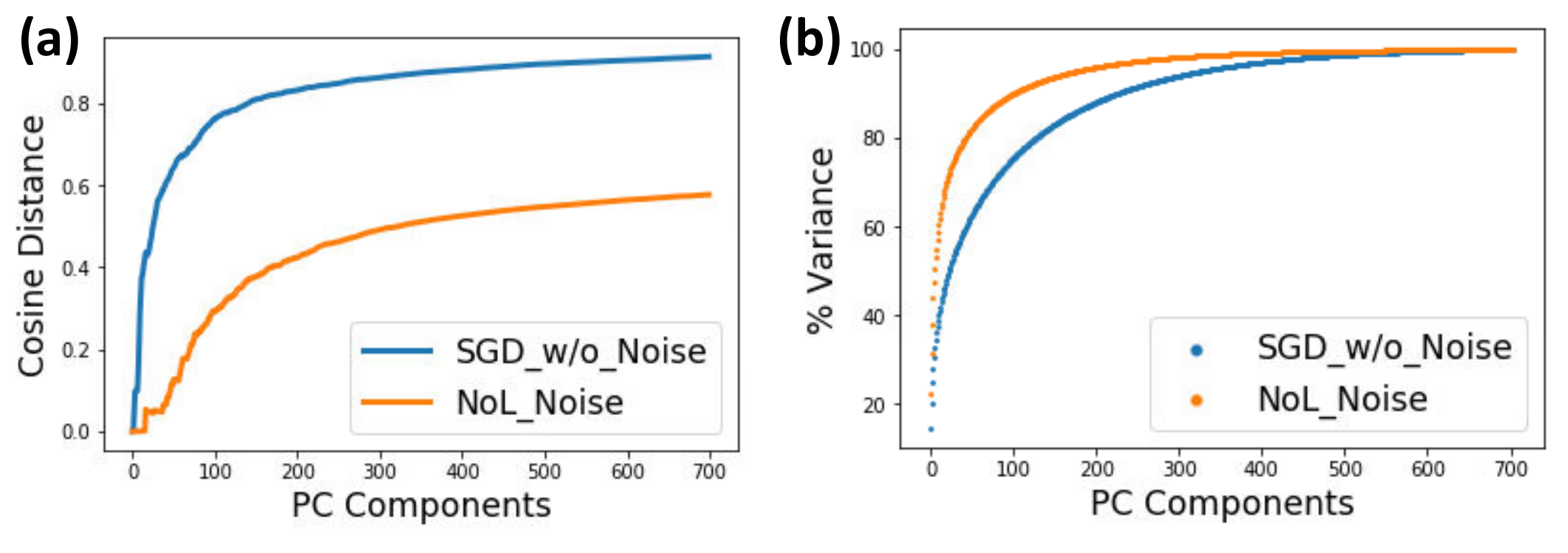}
\caption{(a) Cosine Distance between the model's response to clean and adversarial inputs in the PC subspace. (b) Variance of the $Conv1$ layer of ResNet18 model. (a), (b) compare the SGD/ NoL training scenarios.}
\label{fig3_2}
\end{figure*}

To quantify the dissociation of the adversarial and clean projections in the PC subspace, we calculate the cosine distance ($\textstyle \mathcal{D}^{PC} =\frac{1}{N} \sum_{i=1}^{N}1 - \frac{{\mathcal{F}^{PC}_{clean}}_i \cdot {\mathcal{F}^{PC}_{adv}}_i}{\norm{{\mathcal{F}^{PC}_{clean}}_i}_2 \norm{{\mathcal{F}^{PC}_{adv}}_i}_2}$) between them along different PC dimensions. Here, $N$ represents the total number of sample images used to perform PCA and ${F}^{PC}_{clean} ({F}^{PC}_{adv}$) denote the transformed learnt representations corresponding to clean (adversarial) input, respectively. The distance between the learnt representations (for the \textit{Conv1 layer} of ResNet18 model from the above scenario) consistently increases for latter PCs as shown in Fig. \ref{fig3_2} (a). Interestingly, the cosine distance between adversarial and clean features measured for a model trained with NoL noise is significantly lesser than a standard SGD trained model. This indicates that noise enables the model to look in the vicinity of the original data point and inculcate more adversarial data into its underlying representation. Note, we consider projection across all former dimensions (say, PC0, PC1,...PC100) to calculate the distance at a later dimension (say, PC100) i.e., $
{\mathcal{D}^{PC}}_{100}$ is calculated by taking the dot product between two 100-dimensional vectors: $\mathcal{F}^{PC}_{clean}, \mathcal{F}^{PC}_{adv}$. Please note, in the remainder of the paper, \textit{NoL noise} denotes the prior or noise learnt during the training procedure by minimizing the loss of a neural network. As per Algorithm 1, we use the same set of templates over each training minibatch that ensures that the noise templates randomly initialized at the beginning of training learn and evolve to model the input characteristics (as illustrated in Fig. \ref{fig1}). Specifically, \textit{NoL noise} refers to the learnt noise templates $N: \{N^1,...N^k\}$ as shown in Algorithm 1. 

To further understand the role of NoL noise in a model's behavior, we analyzed the variance captured in the \textit{Conv1} layer's activations of the ResNet18 model (in response to clean inputs) by different PCs, as illustrated Fig. \ref{fig3_2} (b). If $s_{i}=\{1,...,M\}$ are the singular values of the matrix $S$, the variance along a particular dimension $PC_k$ is defined as: $Var_k = 100 \times (\sum_{i=0}^{k}{s_i}^2/ \sum_{i=0}^{M}{s_i}^2)$. $Var_k$ along different PCs provides a good measure of how much a particular dimension explains about the data. We observe that NoL noise increases the variance along the high rank PCs, for instance, the net variance obtained from PC0-PC100 with NoL Noise ($90\%$) is more than that of standard SGD ($76\%$). In fact, we observe a similar increase in variance in the leading PC dimensions for other intermediate blocks’ learnt activations of the ResNet18 model [See \textit{Appendix B}]. We can infer that the increase in variance along the high-rank PCs is a consequence of inclusion of more data points during the overall learning process. Conversely, we can also interpret this as NoL noise embracing more off-manifold adversarial points into the overall data manifold that eventually determines the model's behavior. 
It is worth mentioning that the variance analysis of the model's behavior in response to adversarial inputs yields nearly identical results as Fig. \ref{fig3_2} (b) [\textit{Appendix B}].  

Interestingly, the authors in \citep{hendrycks2017early} conducted PCA whitening of the raw image data for clean and adversarial inputs and demonstrated that adversarial image coefficients for later PCs have greater variance. Our results from PC subspace analysis corroborates their experiments and further enables us to peek into the model's behavior for adversarial attacks. Note, for all the PCA experiments above, we used 700 random images sampled from the CIFAR-10 test data, i.e. $N=700$. In addition, we used the Fast Gradient Sign Method (FGSM) method to create BB adversaries with a step size of $8/255$, from a different source model (ResNet18 trained with SGD). 

\section{Results}
\subsection{Attack Methods}
Given a test image $X$, an attack model perturbs the image to yield an adversarial image, $X_{adv} = X + \Delta$, such that a classifier $f$ misclassifies $X_{adv}$. In this work, we consider $\ell_\infty$ bounded adversaries studied in earlier works \citep{goodfellow2014explaining,tramer2017ensemble,madry2017towards}, wherein the perturbation ($\norm{\Delta}_\infty \le \epsilon$) is regulated by some parameter $\epsilon$. Also, we study robustness against both BB/WB attacks to gauge the effectiveness of our approach.  For an exhaustive assessment, we consider the same attack methods deployed in \cite{tramer2017ensemble,madry2017towards}: 
\\
\textbf{Fast Gradient Sign Method (FGSM):} This single-step attack is a simple way to generate malicious perturbations in the direction of the loss gradient $\nabla_X \mathcal{L}(X, Y_{true})$ as: $X_{adv} = X + \epsilon sign(\nabla_X \mathcal{L}(X, Y_{true}))$.\\
\textbf{Random Step FGSM (R-FGSM):} \citep{tramer2017ensemble} suggested to prepend single-step attacks with a small random step to escape the non-smooth vicinity of a data point that might degrade attacks based on single-step gradient computation. For parameters $\epsilon, \alpha$ ($\alpha=\epsilon/2$), the attack is defined as: $X_{adv} = X’ + \epsilon sign(\nabla_X’ \mathcal{L}(X’, Y_{true})), \hspace{1mm} where \hspace{1mm} X’=X+\alpha sign(\mathcal{N}(0^d, I^d))$.\\
\textbf{Iterative FGSM (I-FGSM):} This method iteratively applies FGSM $k$ times with a step size of $\beta \ge \epsilon/k$ and projects each step perturbation to be bounded by $\epsilon$. Following \citep{tramer2017ensemble}, we use two-step iterative FGSM attacks.\\ 
\textbf{Projected Gradient Descent (PGD):} Similar to I-FGSM, this is a multi-step variant of FGSM:  ${X_{adv}}^{t+1} = \prod({X_{adv}}^{t} + \alpha sign(\nabla_X \mathcal{L}(X, Y_{true})))$ . \cite{madry2017towards} show that this is a universal first-order adversary created by initializing the search for an adversary at a random point followed by several iterations of FGSM. PGD attacks, till date, are one of the strongest BB/ WB adversaries.

\subsection{Experiments}
We evaluated NoL on three datasets: MNIST, CIFAR10 and CIFAR100. For each dataset, we report the accuracy of the models against BB/WB attacks (crafted from the test data) for 6 training scenarios: a) Standard $SGD$ (without noise), b) $NoL$, c) \underline{Ens}emble \underline{Adv}ersarial (EnsAdv) Training ($SGD_{ens}$), d) NoL with EnsAdv Training ($NoL_{ens}$), e) \underline{PGD} \underline{Adv}ersarial (PGDAdv) Training ($SGD_{PGD}$), f) NoL with PGDAdv Training ($NoL_{PGD}$). Note, $SGD_{ens}$ and $SGD_{PGD}$ refer to the standard adversarial training employed in \cite{tramer2017ensemble} and \cite{madry2017towards}, respectively. Our results compare how the additional noise modeling improves over standard SGD in adversarial susceptibility. Also, we integrate NoL with state-of-the-art PGD/Ensemble adversarial training techniques to analyze how noise modeling benefits them.  In case of EnsAdv training, we augmented the training dataset of the target model with adversarial examples (generated using FGSM), from an independently trained model, with same architecture as the target model. In case of PGDAdv training, we augmented the training dataset of the target model with adversarial examples (generated using PGD) from the same target model. Thus, as we see later, EnsAdv imparts robustness against BB attacks only, while, PGD makes a model robust to both BB/WB attacks. In all experiments below, we report the WB/BB accuracy against strong adversaries created with PGD attack. In additon, for BB, we also report the worst-case error over all small-step attacks FGSM, I-FGSM, R-FGSM, denoted as \textit{Min BB} in Table \ref{tab1}, \ref{tab2}. 

Note, $NoL$ scenario refers to our proposed approach (or Algorithm 1) wherein we introduce multiplicative noise at the beginning of training and eventually learn the noise while minimizing the loss for model parameters. $NoL_{ens}$ refers to the scenario wherein we combine EnsAdv training with NoL (or Algorithm 1 with EnsAdv training). That is, a model is trained with adversarial data augmentation and also has multiplicative noise that is eventually learnt during the training procedure. In this case, since we show both clean and FGSM-based BB adversarial input data during training, we can expect that the learnt noise will model both clean/BB-adversarial input distribution. $NoL_{PGD}$ refers to the scenario where PGDAdv training is combined with NoL (or Algorithm 1 with PGDAdv training). Here, we perform the prior noise modeling (with multiplicative noise injected at the beginning of training) while training a network with both clean and PGD-based WB adversarial data. We can expect the learnt noise to model both clean/WB-adversarial input distribution in this case. As we will see later, $NoL_{ens}, NoL_{PGD}$ will serve as good indicators of how integrating the prior noise learning approach with the adversarial defense methods, $SGD_{ens}, SGD_{PGD}$, improves their ability to defend against a larger range of perturbations/attacks.

All networks were trained with mini-batch SGD using a batch size of 64 and momentum of 0.9 (0.5) for CIFAR (MNIST), respectively. For CIFAR10, CIFAR100 we used additional weight decay regularization, $\lambda=5e-4$. Note, for noise modeling, we simply used the negative loss gradients ($\nabla_{N} \mathcal{L} \le 0$) without additional optimization terms. In general, NoL requires slightly more epochs of training to converge to similar accuracy as standard SGD, a result of the additional input noise modeling. Also, NoL models, if not tuned with proper learning rate, have a tendency to overfit. Hence, the learning rate for noise ($\eta_{noise}$) was kept 1-2 orders of magnitude lesser than the overall network learning rate ($\eta$) throughout the training process. All networks were implemented in PyTorch.\footnote{\label{note4}\textit{Appendix C} provides a detailed table of different hyperparameters used to train the source and target models in each scenario corresponding to all experiments of Table \ref{tab1}, \ref{tab2} . Appendix C shows different visualization of noise learnt ($N$) in each scenario of Table \ref{tab1}, \ref{tab2}.}
\begin{table*}[htbp]
\caption{\textbf{MNIST Accuracy (in \%) of ConvNet1 target model for different scenarios.} $\epsilon =0.1/0.2/0.3$ for $SGD, NoL, SGD_{ens}, NoL_{ens}$;  $\epsilon =0.3/0.4$ for $SGD_{PGD}, NoL_{PGD}$.  For PGD attack, we report accuracy for 40-/100-step attacks. Accuracy $<5\%$, in most places, have been omitted and marked as `-'.}
\label{tab1}
\centering
\resizebox{0.85\textwidth}{!}{\begin{tabular}{lllllllp{\textwidth}} 
\toprule
Scenario& Clean & Min BB & PGD-40 & PGD-100 &  PGD-40 & PGD-100 \\
& &\multicolumn{3}{c}{(\textbf{---------------------BlackBox---------------})} & \multicolumn{2}{c}{(\textbf{-----WhiteBox-----})}\\ 
\midrule
$SGD$ & 99.1 & 77.9/20.6/4.3 & 75/9.9/- & 74.5/8/- &  22.3/-/-& - \\
$NoL$ & 99.2 & 83.6/30.5/9.6 & 80.5/20.6/- & 80/18/- &  29.4/-/- & - \\
$SGD_{ens}$ & 99 & 98.5/92.6/73.2 & 98/89.3/71 & 98.1/88/57 &  2.1/-/- & - \\
$NoL_{ens}$ & 99.1 & 99/94.7/76 & 98.8/93.4/79 & 98.7/91.9/66&  3.3/-/- & -  \\
$SGD_{PGD}$ & 97.9 & 91.8/29 & 93.6/48.7 & 92.3/20 & 90/27 & 86.5/4.5 \\
$NoL_{PGD}$ & 98 & 93/42.2 & 94/60.4 & 92.6/28.7 & 90.7/55.7  & 88/20.1  \\
\bottomrule
\end{tabular}}
\end{table*}

\textbf{MNIST:} For MNIST, we consider a simple network with 2 Convolutional (C) layers with 32, 64 filters, each followed by 2$\times$2 Max-pooling (M), and finally a Fully-Connected (FC) layer of size 1024, as the target model (ConvNet1: 32C-M-64C-M-1024FC). We trained 6 ConvNet1 models independently 
corresponding to the different scenarios. 
The EnsAdv ($NoL_{ens}, SGD_{ens}$) models were trained with BB adversaries created from a separate SGD-trained ConvNet1 model using FGSM with $\epsilon=0.1$. PGDAdv ($NoL_{PGD}, SGD_{PGD}$) models were trained with WB adversaries created from the same target model using PGD with $\epsilon=0.3$, step-size = $0.01$ over 40 steps. 

Table \ref{tab1} (Columns 3 - 5) illustrates our results for BB attacks under different perturbations ($\epsilon$)\footnote{\label{note3}For fair comparison, BB attacks on $SGD, NoL, SGD_{ens}, NoL_{ens}$ were crafted from another model trained with standard SGD on natural examples as in \citep{tramer2017ensemble}. While, BB attacks on $SGD_{PGD}, NoL_{PGD}$ were crafted from a model trained with PGDAdv training (without noise modeling) on adversarial examples as in \citep{madry2017towards} to cast stronger attacks.}. NoL noise considerably improves the robustness of a model toward BB attacks compared to standard SGD. Without adversarial training, standard $SGD$ and $NoL$ suffer a drastic accuracy decline for $\epsilon >0.1$. With EnsAdv training, the robustness improves across all $\epsilon$ for both $SGD_{ens}, NoL_{ens}$. However, the robustness is significantly accentuated in $NoL_{ens}$ owing to the additional noise modeling performed during EnsAdv training. An interesting observation here is that for $\epsilon=0.1$ (that was the perturbation size for EnsAdv training), both $NoL_{ens}/ SGD_{ens}$ yield nearly similar accuracy, $\sim98\%$. However, for larger perturbation size $\epsilon=0.2, 0.3$, the network adversarially trained with NoL noise shows higher prediction capability ($\sim>5\%$) across the PGD attack methods. Columns 6-7 in Table \ref{tab1} show the WB attack results. All techniques except for the ones with PGDAdv training fail miserably against the strong WB PGD attacks. Models trained with NoL noise, although yielding low accuracy, still perform better than SGD. $NoL_{PGD}$ yields better accuracy than $SGD_{PGD}$even beyond what the network is adversarially trained for ($\epsilon > 0.3$). Note, for PGD attack in Table \ref{tab1}, we used a step-size of $0.01$ over 40/100 steps to create adversaries bounded by $\epsilon =0.1/0.2/0.3$.  
We also evaluated the worst-case accuracy over all the BB attack methods when the source model is trained with NoL noise (not shown). We found higher accuracies in this case, implying NoL models \textit{transfer} attacks at lower rates. As a result, in the remainder of the paper, we conduct BB attacks from models trained without noise modeling to evaluate the adversarial robustness.

\textbf{CIFAR:} For CIFAR10, we examined our approach on the ResNet18 architecture. We used the ResNext29(2$\times$64d) architecture \citep{xie2017aggregated} with bottleneck width 64, cardinality 2 for CIFAR100. Similar to MNIST, we trained the target models separately corresponding to each scenario and crafted BB/WB attacks. 
For EnsAdv training, we used BB adversaries created using FGSM ($\epsilon=8/255$) from a separate SGD-trained network different from the BB source/target model. For PGDAdv training, the target models were trained with WB adversaries created with PGD with $\epsilon=8/255$, step-size=$2/255$ over 7 steps. Here, for PGD attacks, we use 7/20 steps of size $2/255$ bounded by $\epsilon$. The results appear in Table \ref{tab2}. 

For BB, we observe that $NoL$ ($81\%/63.2\%$ for CIFAR10/100) significantly boosts the robustness of a model as compared to $SGD$ ($50.3\%/44.2\%$ for CIFAR10/100). Note, the improvement here is quite large in comparison to MNIST (that shows only $5\%$ increase from $SGD$ to $NoL$). In fact, the accuracy obtained with $NoL$ alone with BB attack, is almost comparable to that of an EnsAdv/PGDAdv trained model without noise ($SGD_{ens}, SGD_{PGD}$). The richness of the data manifold and feature representation space for larger models and complex datasets allows NoL to model better characteristics in the noise causing increased robustness. As seen earlier, NoL noise ($NoL_{ens}, NoL_{PGD}$) considerably improves the accuracy even for perturbations ($\epsilon=(16,32)/255$) greater than what the network is adversarially trained for. The increased susceptibility of $SGD_{ens}, SGD_{PGD}$ for larger $\epsilon$ establishes that its’ capability is limited by the diversity of adversarial examples shown during training. For WB attacks as well, $NoL_{PGD}$ show higher resistance. Interestingly, while $SGD, SGD_{ens}$ yield infinitesimal performance ($< 5\%$), $NoL, NoL_{ens}$ yield reasonably higher accuracy ($>25\%$) against WB attacks. This further establishes the potential of noise modeling in enabling adversarial security. 
It is worth mentioning that BB accuracy of $SGD_{PGD}, NoL_{PGD}$ models in Table \ref{tab1}, \ref{tab2} are lower than $SGD_{ens}, NoL_{ens}$, since the former is attacked with stronger attacks crafted from models trained with PGDAdv\textsuperscript{\ref{note3}}. Attacking the former with similar adversaries as latter yields higher accuracy.
\begin{table*}[htbp]
\setlength{\tabcolsep}{4pt} 
\caption{\textbf{CIFAR10/ CIFAR100 Accuracy (in \%) of ResNet18/ ResNext-29 target model for different scenarios.} $\epsilon = \frac{8}{255}/\frac{16}{255}/\frac{32}{255}$ for $NoL, SGD, NoL_{ens}, SGD_{ens}, NoL_{PGD}, SGD_{PGD}$. For PGD attack, we report accuracy for 7-/20-step attacks. Accuracy $<5\%$, in most places, have been omitted and marked as `-'.}
\label{tab2}
\centering
\resizebox{0.85\textwidth}{!}{\begin{tabular}{lllllllp{\textwidth}}
\toprule
Scenario& Clean & Min BB & PGD-7 & PGD-20 &  PGD-7 & PGD-20 \\
& &\multicolumn{3}{c}{(\textbf{---------------------BlackBox---------------})} & \multicolumn{2}{c}{(\textbf{---------WhiteBox--------})}\\
\midrule
\multicolumn{7}{c}{\textbf{ResNet18 (CIFAR10)}}\\
$SGD$ & 88.8 & 50.3/32/16.2 & 16.2/15/11.8 & 7.5/6.2/- &  - & - \\
$NoL$ & 87.1 & 81/76/67  &  72.6/71.6/69.1 & 37/36.1/33.2 &  8.8/8.2/7.1 & - \\
$SGD_{ens}$ & 86.3 & 81.3/76.6/68.3 & 80.9/75.1/67.1 & 80.2/74.4/63 & 0.8/-/- & - \\
$NoL_{ens}$ & 86.4 & 84.4/81.4/72.6  & 83/80/71.3 & 82.7/79/71 & 29/21/16.5 & -  \\
$SGD_{PGD}$ & 83.2 & 71.3/58/50 & 69.9/62/50.1 & 54.2/50.3/46 & 58.4/48/42 & 57.3/42.8/28 \\
$NoL_{PGD}$ & 83 & 73/62/56.8 & 71/65/53 & 57.6/53.7/49.8 &  63/59/57 & 59.2/45/30.1 \\
\midrule
\multicolumn{7}{c}{\textbf{ResNext29 (CIFAR100)}}\\
$SGD$ & 71 & 44.2/38.4/26.7 & 42.7/35/25.4 & 40.5/27/17 &  - & - \\
$NoL$ & 69.4  & 63.2/58.5/50.1 & 62.9/54.3/48.4 & 62.3/53.1/42.5 &  19/14/10.3 & - \\
$SGD_{ens}$ & 69.8 & 64.8/60.9/50 & 63.6/57.5/45.4 & 63/56/42  & 2.5/-/-  & - \\
$NoL_{ens}$ & 67.3 & 65.1/62.8/57  & 64.8/61.4/52.2 & 64.4/58/49 &  18/14/11 & -  \\
$SGD_{PGD}$ & 71.6 & 57.5/48/38.4  & 56/45/41.3 & 48/40/38.4 & 51.5/49.8/46 & 50.4/43/33 \\
$NoL_{PGD}$ & 69 & 66.3/62/59.9 & 63/58.7/54.1 & 52.3/50/40.8 & 58.1/56/53  & 53/48/37.9 \\
\bottomrule
\end{tabular}}
\end{table*}

\begin{figure*}[h]
\centering
\includegraphics[width=\textwidth]{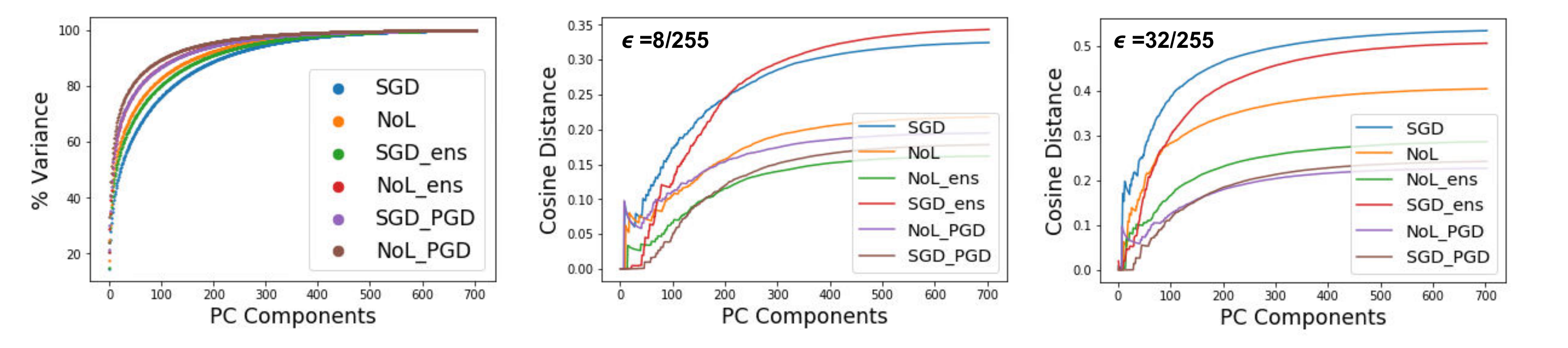}
\caption{[Left] Variance (in response to clean inputs) across different scenarios for the first 700 PC dimensions.  [Middle, Right] Cosine distance across 700 PCs between clean and adversarial representations for varying $\epsilon$. \textbf{$SGD_{ens}, SGD_{PGD}$ exhibit improved variance (and lower distance) than $SGD$, suggesting PC variance/ distance as a good indicator of adversarial robustness.} PCA was conducted with sample of 700 test images.}
\label{fig4_1}
\end{figure*}

\textbf{PC Distance \& Variance Analysis :} Next, we measured the variance and cosine distance captured by the $Conv1$ layer of the ResNet18 model corresponding to different scenarios (Table \ref{tab2}). 
Fig. \ref{fig4_1} shows that variance across the leading PCs decreases as $NoL_{PGD}>SGD_{PGD}>NoL_{ens}>NoL>SGD_{ens}>SGD$.  Inclusion of adversarial data points with adversarial training or noise modeling informs a model more, leading to improved variance. We note that $NoL_{ens}$ and $SGD_{PGD}$ yield nearly similar variance ratio, although $SGD_{PGD}$ gives better accuracy than $NoL_{ens}$ for similar BB and WB attacks. Since we are analyzing only the $Conv1$ layer, we get this discrepancy. 
In Fig. \ref{fig4_1}, we also plot the cosine distance between the adversarial (created from FGSM with specified $\epsilon$) and clean inputs in the PC subspace. The distance across different scenarios along latter PCs increases as: $NoL_{PGD}<SGD_{PGD}<NoL_{ens}<NoL<SGD_{ens}<SGD$. A noteworthy observation here is, PC distance follows the same order as decreasing variance and justifies the accuracy results in Table \ref{tab2}. The decreasing distance with $NoL$ compared to $SGD$ further signifies improved realization of the on-/off-manifold data. Also, the fact that $NoL_{PGD}, NoL_{ens}$ have lower distance for varying $\epsilon$ establishes that integrating noise modeling with adversarial training compounds adversarial robustness. Interestingly, for both variance and PC distance, $NoL$ has a better characteristic than $SGD_{ens}$. This proves that noise modeling enables implicit inclusion of adversarial data without direct data augmentation, as opposed to EnsAdv training (or $SGD_{ens}$) where the dataset is explicitly augmented. This also explains the comparable BB accuracy between $NoL, SGD_{ens}$ in Table \ref{tab2}. 

\begin{figure*}[h]
\centering
\includegraphics[width=\textwidth]{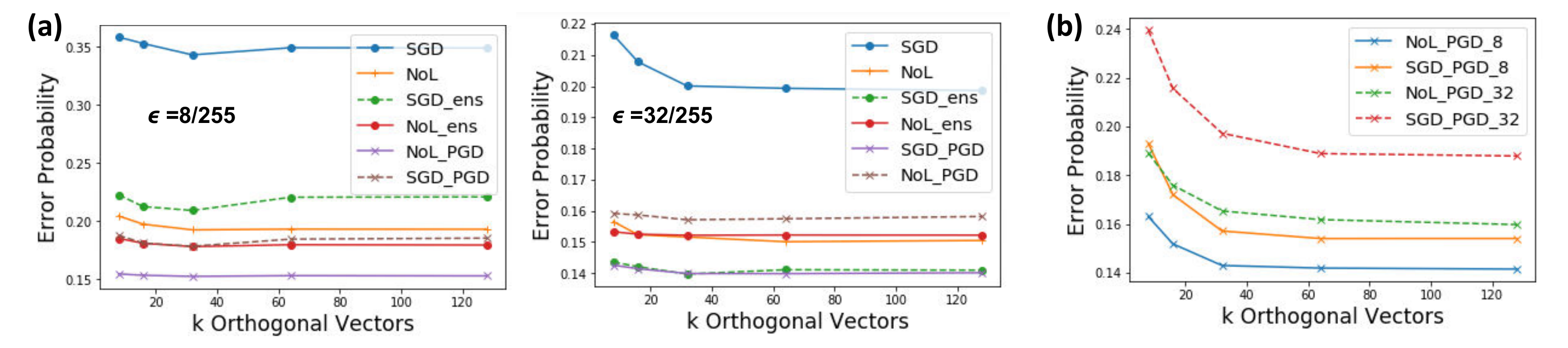}
\caption{Adversarial subspace dimensionality for varying $\epsilon$ for- (a) -BB adversaries crafted from a model trained with natural examples (b) -WB adversaries crafted for models trained with PGDAdv training. }
\label{fig4_2}
\end{figure*}

 \textbf{Adversarial Subspace Dimensionality :} To further corroborate that NoL noise implicitly embraces adversarial points, we evaluated the adversarial subspace dimension using the Gradient-Aligned Adversarial Subspace (GAAS) method of \citep{tramer2017space}. We construct $k$ orthogonal vectors $r_1,..,r_k \in \{-1,1\}$ from a regular Hadamard matrix of order $k \in \{2^2, 2^3,..,2^7\}$. We then multiply each $r_i$ component-wise with the gradient, $sign(\nabla_X\mathcal{L}(X,Y_{true}))$. Hence, estimating the dimensionality reduces to finding a set of orthogonal perturbations, $r_i$ with $\norm{r_i}_\infty = \epsilon$ in the vicinity of a data point that causes misclassification. For each scenario of Table \ref{tab2} (CIFAR10), we select 350 random test points, $x$, and plot the probability that we find at least $k$ orthogonal vectors $r_i$ such that $x+r_i$ is misclassified. Fig. \ref{fig4_2} (a), (b) shows the results with varying $\epsilon$ for BB, WB instances. We find that the size of the space of adversarial samples is much lower for a model trained with NoL noise than that of standard SGD. For $\epsilon =8/255$, we find over 128/64 directions for $\sim25\%/15\%$ of the points in case of $SGD/NoL$. With EnsAdv training, the number of adversarial directions for $SGD_{ens}/NoL_{ens}$ reduces to 64 that misclassifies $\sim17/15\%$ of the points. With PGDAdv training, the adversarial dimension significantly reduces in case of $NoL_{PGD}$ for both BB/WB. As we increase the perturbation size ($\epsilon =32/255$), we observe increasingly reduced number of misclassified points as well as adversarial dimensions for models trained with noise modeling. The WB adversarial plot, in Fig. \ref{fig4_2} (b), clearly shows the reduced space obtained with noise modeling with PGDAdv training ($NoL_{PGD}$) against plain PGDAdv ($SGD_{PGD}$) for $\epsilon =(8, 32)/255$. 
\\

\begin{figure}[ht]
\centering
\includegraphics[width=0.5\textwidth]{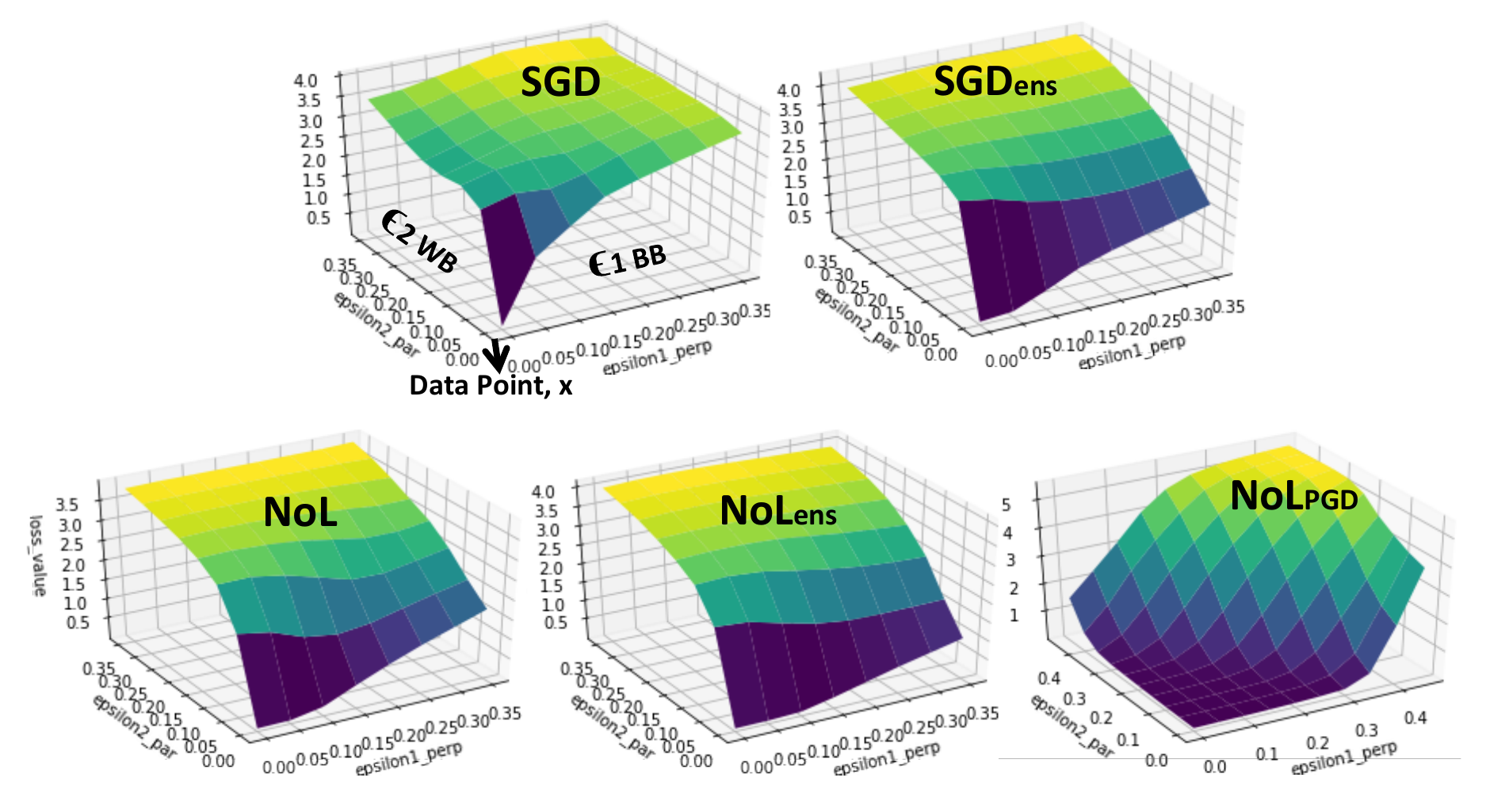}
\caption{Loss surface of models corresponding to MNIST (Table\ref{tab1}).}
\label{fig5}
\end{figure}

\textbf{Loss Surface Smoothening:} By now, it is clear that while NoL alone can defend against BB attacks (as compared to SGD) reasonably well, it still remains vulnerable to WB attacks. For WB defense and to further improve BB defense, we need to combine NoL noise modeling with adversarial training. To further investigate this, we plotted the loss surface of MNIST models on examples $x’ = x+\epsilon_1 \cdot g_{BB} +\epsilon_2 \cdot g_{WB}$ in Fig. \ref{fig5}, where $g_{BB}$ is the signed gradient, $sign({\nabla_X\mathcal{L}(X,Y_{true})}_{source})$, obtained from the source model (crafting the BB attacks) and $g_{WB}$ is the gradient obtained from the target model itself (crafting WB attacks), $sign({\nabla_X\mathcal{L}(X,Y_{true})}_{target})$. We see that the loss surface in case of $SGD$ is highly curved with steep slopes in the vicinity of the data point in both BB and WB direction. The EnsAdv training, $SGD_{ens}$, smoothens out the slope in the BB direction substantially, justifying their robustness against BB attacks. Models trained with noise modeling, $NoL$ (even without any data augmentation), yield a softer loss surface. This is why $NoL$ models \textit{transfer} BB attacks at lower rates. The surface in the WB direction along $\epsilon_2$ with $NoL, NoL_{ens}$ still exhibits a sharper curvature (although slightly softer than $SGD_{ens}$) validating the lower accuracies against WB attacks (compared to BB attacks). PGDAdv, on the other hand, smoothens out the loss surface substantially in both directions owing to the explicit inclusion of WB adversaries during training. Note, $NoL_{PGD}$ yields a slightly softer surface than $SGD_{PGD}$ (not shown). The smoothening effect of noise modeling further justifies the boosted robustness of NoL models for larger perturbations (outside $\epsilon$-ball used during adversarial training).   
It is worth mentioning that we get similar PCA/ Adversarial dimensionality/ loss surface results across all datasets.
\section{Discussion}
We proposed \textit{Noise-based Prior Learning, NoL,} as a reliable method for improving adversarial robustness. Specifically, our key findings are:\\
1) \hspace{5pt} We show that noise modeling at the input during discriminative training improves a model's ability to generalize better for out-of-sample adversarial data (without explicit data augmentation).\\
2) \hspace{5pt} Our PCA variance and cosine distance analysis provides a significant perspective to visualize and quantify a model's response to adversarial/clean data. 

A crucial question one can ask is, \textbf{\textit{How to break NoL defense?}} The recent work \citep{athalye2018obfuscated} shows that many defense methods cause `gradient masking' that eventually fail. We reiterate that, NoL alone does not give a strong BB/WB defense. However, the smoothening effect of noise modeling on the loss (Fig. \ref{fig5}) suggests that noise modeling decreases the magnitude of the gradient masking effect. NoL does not change the classification model that makes it easy to be scaled to larger datasets while integrating with other adversarial defense techniques. Coupled with other defense, NoL performs remarkably (even for larger $\epsilon$ values). We combine NoL with EnsAdv \& PGDAdv, which do not cause obfuscated gradients and hence can withstand strong attacks, however, upto a certain point. For WB perturbations much greater than the training $\epsilon$ value, NoL+PGDAdv also breaks. In fact, for adaptive BB adversaries \cite{tramer2017ensemble} or adversaries that query the model to yield full prediction confidence (not just the label), NoL+EnsAdv will be vulnerable.  Note, advantage with NoL is, being independent of the attack/defense method, NoL can be potentially combined with stronger attacks developed in future, to create stronger defenses.

While variance and principal subspace analysis help us understand a model's behavior, we cannot fully describe the structure of the manifold learnt by the linear subspace view. 
However, PCA does provide a basic intuition about the generalization capability of complex image models. In fact, our PC results establish the superiority of PGDAdv training (\citep{ madry2017towards} best defense so far), in general, as a strong defense method and can be used as a valid metric to gauge adversarial susceptibility in future proposals. Finally, as our likelihood theory indicates, better noise modeling techniques with improved gradient penalties can further improve robustness and requires further investigation. Also, performing noise modeling at intermediate layers to improve variance, and hence robustness, are other future work directions.

\textbf{\textit{A final note on novelty of NoL.}} As noted earlier, there have been past works \cite{noh2017regularizing, fawzi2016robustness, HintonLec} where the training dataset is perturbed with random noise, and the perturbed images are then used to train a deep neural network. The role of such noise is to impose better regularization effect (or reduce overfitting) during training which, in turn, improves a model’s generalization capability or inference accuracy \cite{HintonLec}. It is worth mentioning that such methods do not result in adversarial robustness (also, verified in \cite{goodfellow2014explaining}). In the proposed $NoL$ approach, the input noise (randomly initialized at the beginning of training) is gradually learnt during the training procedure. The fact that we train the noise (or prior) is the major distinguishing factor between our approach and previous works \cite{noh2017regularizing, fawzi2016robustness, HintonLec} that simply perturb the training data with random noise. The implicit prior modeling property of NoL goes beyond simple regularization and in fact, results in adversarial robustness (wherein, a model's ability to generalize on adversarial data improves). Our PC distance and variance analysis (in Section 2.3, 3.2) further attest the capability of the learnt noise (or prior) to embrace adversarial points into the data manifold thereby improving a model's adversarial accuracy. In fact, our results in Table \ref{tab2} show that standalone $NoL$ achieves better white-box accuracy than the ensemble adversarial training defense approach $SGD_{ens}$. This implies that even some white-box perturbations are inherited in this implicit noise modeling procedure. Such adversarial robustness cannot be attained with a setup of training over randomly perturbed examples as \cite{noh2017regularizing, fawzi2016robustness}. Finally, we would also like to point out that our noise-based prior learning can possibly render some regularization effects on a model during training. The smoothened loss curves in Fig. \ref{fig5} observed with $NoL$ in comparison to $SGD$ verifies the regularized model behavior. 

\section*{Acknowledgments}
This work was supported in part by, Center for Brain-inspired Computing Enabling Autonomous Intelligence
(C-BRIC), a DARPA sponsored JUMP center, by the Semiconductor Research Corporation, the National Science Foundation, Intel Corporation, the DoD Vannevar Bush Fellowship and by the U.S. Army Research Laboratory and the U.K. Ministry of Defense under Agreement Number W911NF-16-3-0001.

\newpage

\renewcommand{\thefigure}{A\arabic{figure}}
\renewcommand{\thetable}{A\arabic{table}}
\setcounter{figure}{0}    
\setcounter{table}{0}    
\setcounter{section}{0}    
\renewcommand\thesection{\Alph{section}}
\renewcommand\thesubsection{\thesection.\arabic{subsection}}

\section{Appendix A: Justification of $X+N$ vs $X \times N$ and use of $\nabla \mathcal{L}_{N} \le 0$ for noise modeling}
\begin{figure}[h]
\centering
\includegraphics[width = \linewidth]{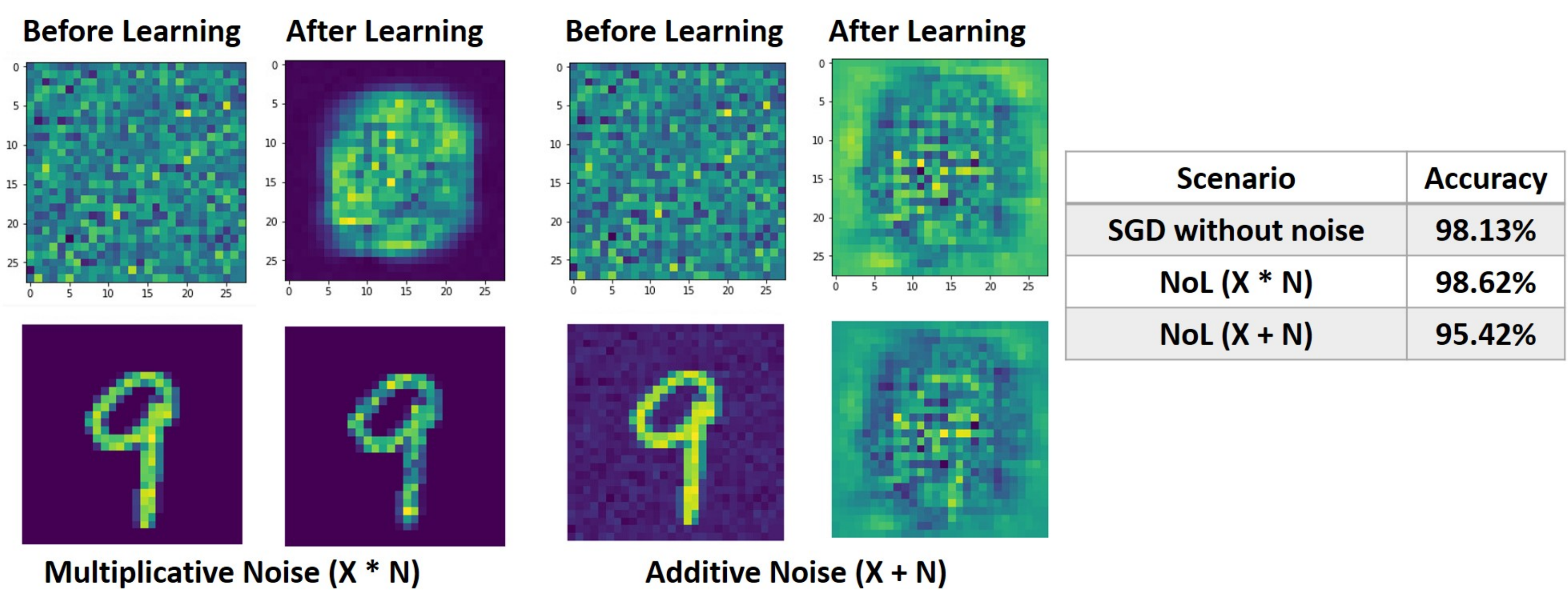}
\caption{\label{figs1} For MNIST dataset, we show the noise template learnt when we use multiplicative/additive noise ($N$) for Noise-based Prior Learning. The final noise-integrated image (for a sample digit `9') that is fed to the network before and after training is also shown. Additive noise disrupts the image drastically. Multiplicative noise, on the other hand, enhances the relevant pixels while eliminating the background. Accuracy corrsponding to each scenario is also shown and compared against standard SGD training scenario (without any noise). Here, we train a simple convolutional architecture (ConvNet: 10C-M-20C-M-320FC) of 2 Convolutional (C) layers with 10, 20 filters, each followed by 2$\times$2 Max-pooling (M) and a Fully-Connected (FC) layer of size 320. We use mini-batch SGD with momentum of 0.5, learning rate ($\eta$=0.1) decayed by 0.1 every 15 epochs and batch-size 64 to learn the network parameters. We trained 3 ConvNet models independently corresponding to each scenario for 30 epochs. For the NoL scenarios, we conduct noise modelling with only negative loss gradients ($\nabla \mathcal{L}_{N} \le 0$) with noise learning rate, $\eta_{noise} = 0.001$, throughout the training process. Note, the noise image shown is the average across all 64 noise templates.}
\vspace{4mm}
\end{figure}

\newpage

\begin{figure}[h]
\centering
\includegraphics[width = \linewidth]{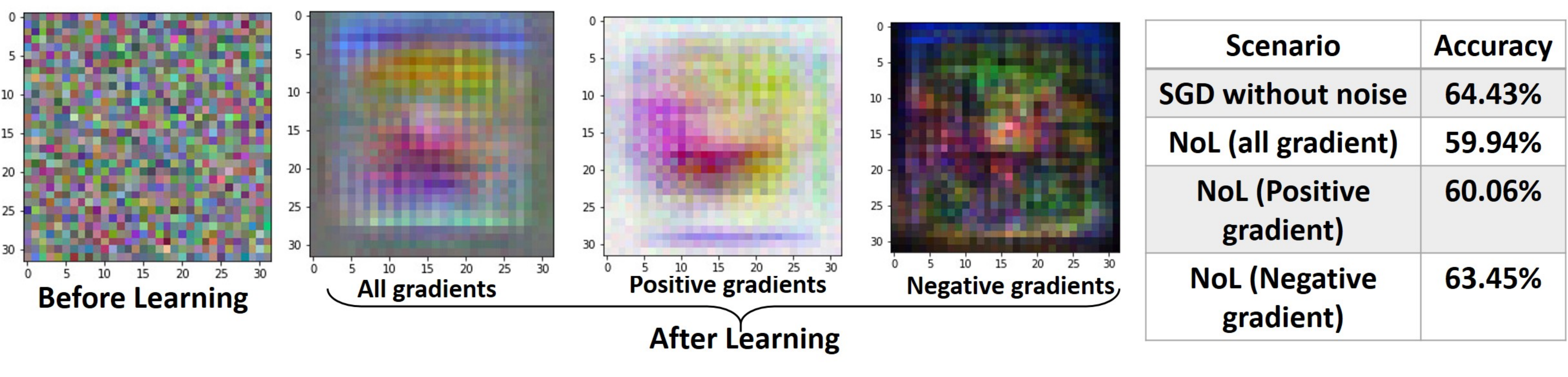}
\caption{\label{figs2} Here, we showcase the noise learnt by a simple convolutional network (ConvNet: 10C-M-20C-M-320FC), learning the CIFAR10 data with NoL (multiplicative noise) under different gradient update conditions. As with MNIST (Fig. \ref{figs1}), we observe that the noise learnt enhances the region of interest while deemphasizing the background pixels. Note, the noise in this case has RGB components as a result of which we see some prominent color blobs in the noise template after training. The performance table shows that using only negative gradients (i.e. $\nabla \mathcal{L}_{N} \le 0$) during backpropagation for noise modelling yields minimal loss in accuracy as compared to a standard SGD trained model.  We use mini-batch SGD with momentum of 0.9, weight decay 5e-4, learning rate ($\eta$=0.01) decayed by 0.2 every 10 epochs and batch-size 64 to learn the network parameters. We trained 4 ConvNet models independently corresponding to each scenario for 30 epochs. For the NoL scenarios, we conduct noise modelling by backpropagating the corresponding gradient with noise learning rate ($\eta_{noise} = 0.001$) throughout the training process. Note, the noise image shown is the average across all 64 noise templates. }
\vspace{4mm}
\end{figure}

\newpage
\section{Appendix B: PC variance for $SGD$ and $NoL$ scenarios in response to adversarial and clean inputs across different layers of ResNet18}
\begin{figure}[h]
\centering
\includegraphics[width = \linewidth]{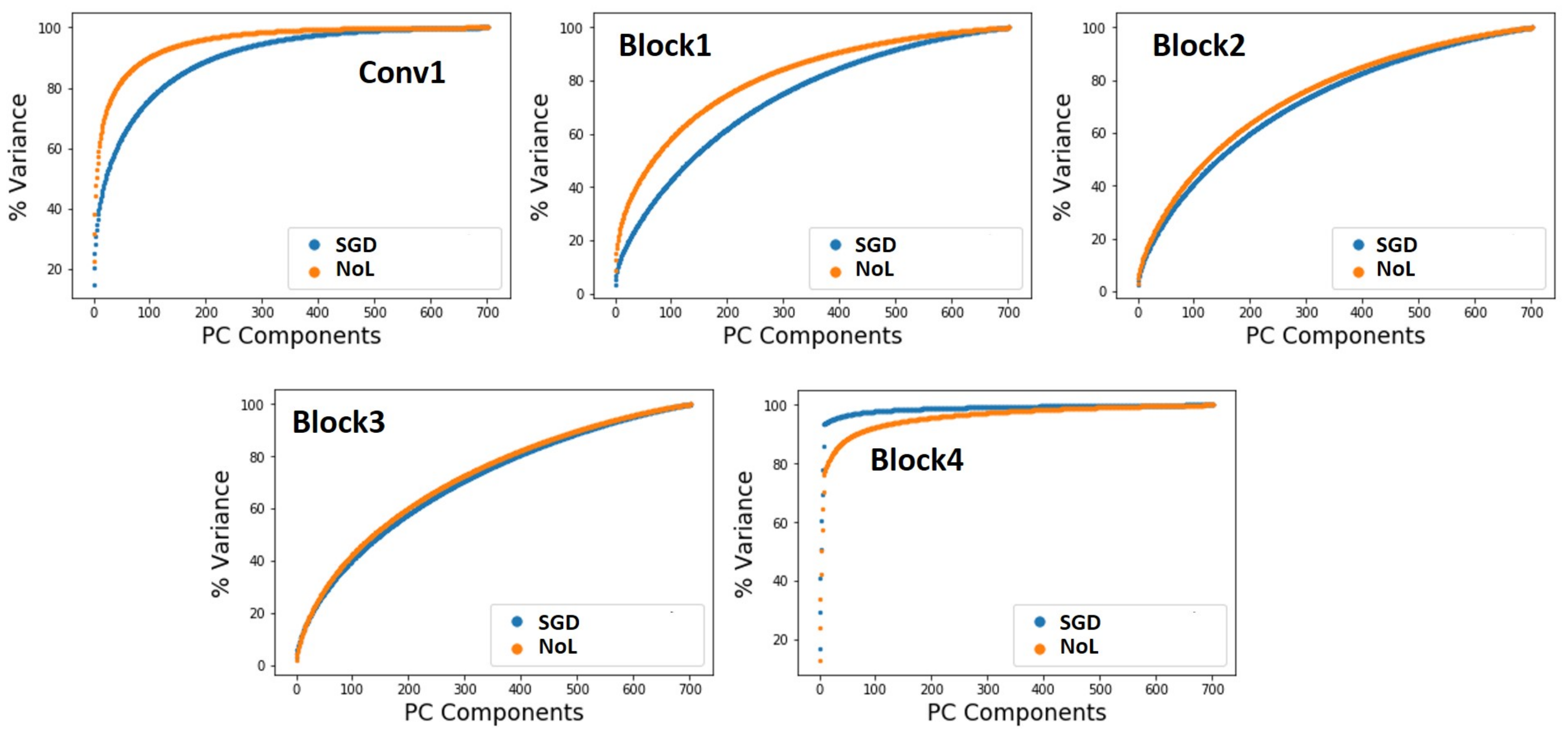}
\caption{\label{figs3} Here, we show the variance captured in the leading Principal Component (PC) dimensions for the inital convolutional layer's ($Conv1$) and intermediate blocks’ learnt activations of a ResNet-18 model trained on CIFAR10 data. We compare the variance of the learnt representations (in response to clean inputs) for each block across two scenarios: SGD (without noise) and NoL (with noise). Note, we capture the variance of the final block's activations before average pooling. That is, the activations of $Block4$ have dimension $512\times4\times4$. We observe that NoL noise increases the variance along the high rank PCs. Also, as we go deeper into the network, the absolute difference of the variance values between $SGD/NoL$ decreases. This is expected as the contribution of input noise on the overall representations decreases as we go deeper into the network. Moreover, there is a generic-to-specific transition in the hierarchy of learnt features of a deep neural network. Thus, the linear PC subspace analysis to quantify a model's knowledge of the data manifold is more applicable in the earlier layers, since they learn more general input-related characteristics. Nonetheless, we see that NoL model yields widened variance than $SGD$ for each intermediate layer except the final $Block4$ that feeds into the output layer.  We use mini-batch SGD with momentum of 0.9, weight decay 5e-4, learning rate ($\eta$=0.1) decayed by 0.1 every 30 epochs and batch-size 64 to learn the network parameters. We trained 2 ResNet-18 models independently corresponding to each scenario for 60 epochs. For noise modelling, we use $\eta_{noise} = 0.001$ decayed by 0.1 every 30 epochs. Note, we used a sample set of 700 test images to conduct the PCA.}
\end{figure}

\pagebreak
\begin{figure}[h]
\centering
\includegraphics[width = \linewidth]{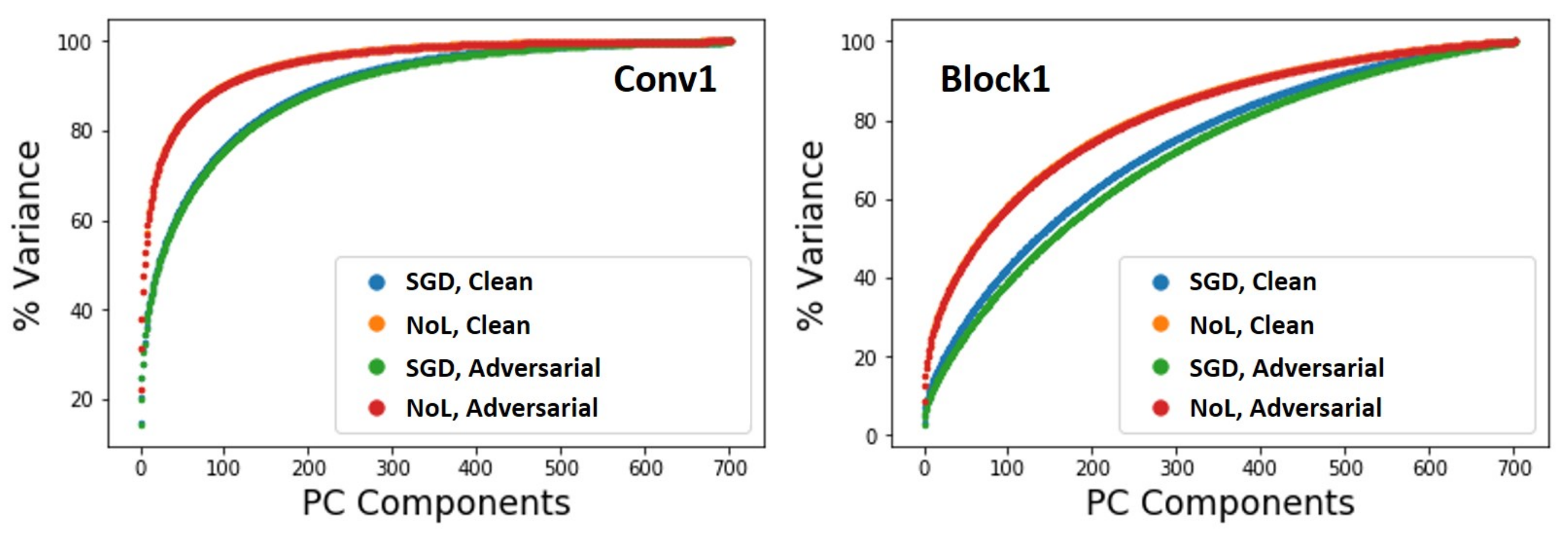}
\caption{\label{figs4} Here, we show the variance captured in the leading Principal Component (PC) dimensions for the $Conv1$ and $Block1$ learnt activations in response to both clean and adversarial inputs for ResNet-18 models correponding to the scenarios discussed in Fig. \ref{figs3}. The model's variance for both clean and adversarial inputs are exactly same in case of $NoL/SGD$ for $Conv1$ layers. For $Block1$, the adversarial input variance is slighlty lower in case of $SGD$ than that of clean input. With $NoL$, the variance is still the same for $Block1$. This indicates that PC variance statistics cannot differentiate between a model's knowledge of on-/off- manifold data. It only tells us whether a model's underlying representation has acquired more knowledge about the data manifold. To analyze a model's understanding of adversarial data, we need to look into the relationship between the clean and adversarial projection onto the PC subspace and measure the cosine distance. Note, we used the Fast Gradient Sign Method (FGSM) method \cite{goodfellow2014explaining} to create BB adversaries with a step size of $8/255$, from another independently trained ResNet-18 model ($source$) with standard SGD. The $source$ attack model has the same hyperparameters as the  $SGD$ model in Fig. \ref{figs3} and is trained for 40 epochs.}
\end{figure}

\pagebreak
\section{Appendix C: Experimental Details and Model Description}
\begin{figure}[h]
\centering
\includegraphics[width = 0.7\linewidth]{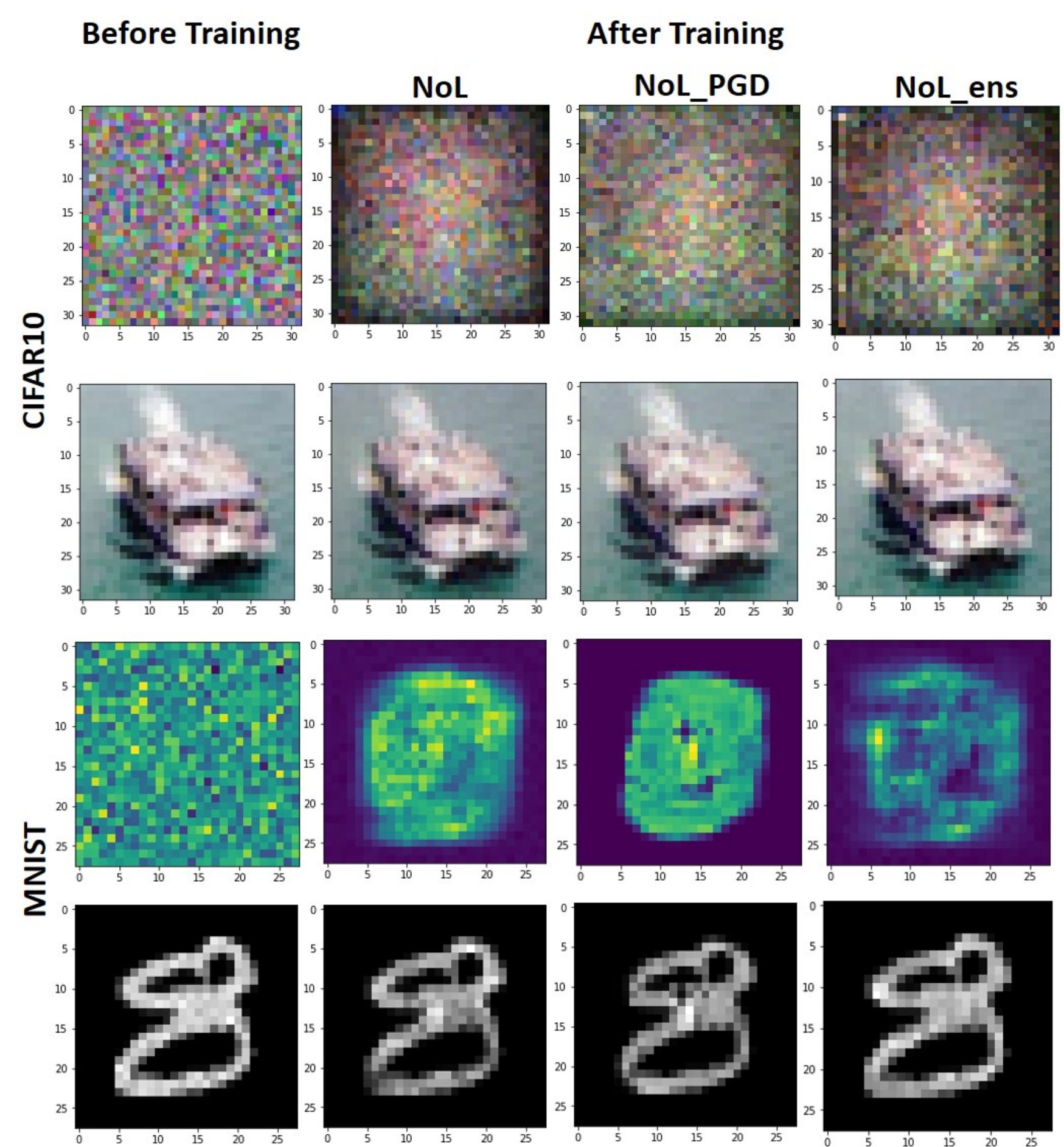}
\caption{\label{figs5} Here, we show the noise templates learnt with noise modeling corresponding to different training scenarios of Table 1, 2 in main paper: NoL (only noise modeling), NoL\_PGD (noise modeling with PGDAdv training $NoL_{PGD}$), NoL\_ens (noise modeling with EnsAdv training $NoL_{ens}$) for MNIST and CIFAR10 data. A sample image ($X \times N$) before and after training with different scenarios is shown. The fact that every training technique yields different noise template shows that noise influences the overall optimization. Column 1 shows the noise template and correponding image ($X \times N$ ) before training, Coulmns 2-4 show the templates after training. Note, noise shown is the mean across 64 templates.}
\end{figure}

The Pytorch implementation of ResNet-18 architecture for CIFAR10 and ResNext-29 architecture for CIFAR100 were taken from \citep{WinNT}. For CIFAR10/CIFAR100, we use mini-batch SGD with momentum of 0.9, weight decay 5e-4 and batch size 64 for training the weight parameters of the models. A detailed description of the learning rate and epochs for ResNet18 model (corresponding to Table 2 in main paper) is shown in Table \ref{tabs3}. Similarly, Table \ref{tabs4} shows the parameters for ResNext-29 model. The hyperparmeters corresponding to each scenario (of Table \ref{tabs3}, \ref{tabs4}) are shown in Rows1-6 under \textit{Target} type. The hyperparameters for the source model used to attack the target models for BB scenarios is shown in Row 7/8 under \textit{Source} type. We use BB attacks from the SGD trained source model to attack $SGD, NoL, NoL_{ens}, PGD_{ens}$. We use BB attacks from a model trained with PGD adversarial training ($\epsilon =8/255$,  \textit{step-size=2/255 over 7 steps}) to craft strong BB attacks on $SGD_{PGD}, NoL_{PGD}$. The model used to generate black box adversaries to augment the training dataset of the $SGD_{ens}, NoL_{ens}$ target models is shown in Row 9 under \textit{EnsAdv} type.  

\textbf{How to conduct Ensemble Adversarial Training?} Furthermore, in all our experiments, for EnsAdv training ($SGD_{ens}$), we use a slightly different approach than \cite{kurakin2016adversarial}. Instead of using a weighted loss function that controls the relative weight of adversarial/clean examples in the overall loss computation, we use a different learning rate $\eta_{adv}/\eta$ ($\eta_{adv}<\eta$) when training with adversarial/clean inputs, respectively, to learn the network parameters. Accordingly, while performing adversarial training with noise-based learning ($NoL_{ens}$), the noise modeling learning rate in addition to overall learning rate, $\eta_{adv}/\eta$, for adversarial/clean inputs is also different, ${\eta_{noise}}_{adv}/{\eta_{noise}}$ (${\eta_{noise}}_{adv}<{\eta_{noise}}$).

\textbf{How to conduct PGD Adversarial Training?} For PGD adversarial training ($SGD_{PGD}$), we used the techniques suggested in \citep{kannan2018adversarial}. \cite{kannan2018adversarial} propose that training on a mixture of clean and adversarial examples (generated using PGD attack), instead of literally solving the min-max problem described by \citep{ madry2017towards} yields better performance. In fact, this helps maintain good accuracy on both clean and adversarial examples. Like EnsAdv training, here as well, we use a different learning rate $\eta_{adv}/\eta$ ($\eta_{adv}<\eta$) when training with adversarial/clean inputs, respectively, to learn the network parameters. Accordingly, while performing PGD adversarial training with noise-based learning ($NoL_{PGD}$), the noise modeling learning rate in addition to overall learning rate, $\eta_{adv}/\eta$, for adversarial/clean inputs is also different, ${\eta_{noise}}_{adv}/{\eta_{noise}}$ (${\eta_{noise}}_{adv}<{\eta_{noise}}$)

Note, the adversarial inputs for EnsAdv training of a target model are created using BB adversaries generated by standard FGSM from a source (shown in Row 9 of Table \ref{tabs3}, \ref{tabs4}), while PGDAdv training uses WB adversaries created with PGD attack from the same target model. We also show the test accuracy (on clean data) for each model in Table \ref{tabs3},\ref{tabs4} for reference. Note, the learning rate in each case decays by a factor of 0.1 every 20/30 epochs (Column 5 in Table \ref{tabs3}, \ref{tabs4}).

\begin{table}[htbp]
\centering
\setlength{\tabcolsep}{1pt} 
\caption{\textbf{Hyperparameter Table for training ResNet18 models on CIFAR10 data}}
\label{tabs3}
\centering
\resizebox{0.5\textwidth}{!}{\begin{tabular}{cccccccc}%
\toprule
Model Type 
&\multicolumn{1}{p{2cm}}{\centering Training \\Method}
& Epochs
& $\eta/ \eta_{adv}$ 
& \multicolumn{1}{p{2cm}}{\centering $\eta, \eta_{adv}$ \\decay/step-size} 
& $\eta_{noise}/{\eta_{noise}}_{adv}$ 
& \multicolumn{1}{p{2cm}}{\centering $\eta_{noise}, {\eta_{noise}}_{adv}$ \\decay/step-size}
& \multicolumn{1}{p{2cm}}{\centering Test \\Accuracy in (\%)}\\
\midrule
\multirow{6}{4em}{Target}&$SGD$ & 120 & 0.1/-- & 0.1/30 & -- & -- & 88.8\\
&$NoL$ & 120 & 0.1/-- & 0.1/30 & 0.001/-- & 0.1/30 & 87.1\\
&$SGD_{ens}$ & 80 & 0.1/0.05 & 0.1/30 & -- & --  & 86.3\\
&$NoL_{ens}$ & 120 & 0.1/0.05 & 0.1/30 & 0.001/0.0005 & 0.1/30 & 86.4\\
&$SGD_{PGD}$ & 122 & 0.1/0.1 & 0.1/20 & -- & --  & 83.2\\
&$NoL_{PGD}$ & 122 & 0.1/0.1 & 0.1/20 & 0.001/0.0005 & 0.1/20 & 83\\
\midrule
\multirow{2}{4em}{Source}&$SGD$ & 300 & 0.1/-- & 0.1/100 & -- & -- & 89\\
&$PGDAdv$ & 122 & 0.1/0.1 & 0.1/20 & -- & -- & 83\\
\midrule
\multirow{1}{4em}{EnsAdv}&$SGD$ & 31 & 0.1/-- & 0.1/30 & -- & -- & 81\\
\bottomrule
\end{tabular}}
\vspace{3mm}
\end{table}

\begin{table}[htbp]
\centering
\setlength{\tabcolsep}{1pt} 
\caption{\textbf{Hyperparameter Table for training ResNext29 models on CIFAR100 data}}
\label{tabs4}
\centering
\resizebox{0.5\textwidth}{!}{\resizebox{\textwidth}{!}{\begin{tabular}{cccccccc}
\toprule
Model Type 
&\multicolumn{1}{p{2cm}}{\centering Training \\Method}
& Epochs
& $\eta/ \eta_{adv}$ 
& \multicolumn{1}{p{2cm}}{\centering $\eta, \eta_{adv}$ \\decay/step-size} 
& $\eta_{noise}/{\eta_{noise}}_{adv}$ 
& \multicolumn{1}{p{2cm}}{\centering $\eta_{noise}, {\eta_{noise}}_{adv}$ \\decay/step-size}
& \multicolumn{1}{p{2cm}}{\centering Test \\Accuracy in (\%)}\\
\midrule
\multirow{6}{4em}{Target}&$SGD$ & 100 & 0.1/-- & 0.1/40 & -- & -- & 71\\
&$NoL$ & 58 & 0.1/-- & 0.1/20 & 0.001/-- & 0.1/20 & 69.4\\
&$SGD_{ens}$ & 42 & 0.1/0.05 & 0.1/20 & -- & --  & 69.8\\
&$NoL_{ens}$ & 48 & 0.1/0.05 & 0.1/20 & 0.001/0.0005 & 0.1/20 & 67.3\\
&$SGD_{PGD}$ & 52 & 0.1/0.05 & 0.1/20 & -- & --  & 71.6\\
&$NoL_{PGD}$ & 52& 0.1/0.05 & 0.1/20 & 0.001/0.0005 & 0.1/20 & 69\\
\midrule
\multirow{2}{4em}{Source}&$SGD$ & 34 & 0.1/-- & 0.1/10 & -- & -- & 67.2\\
&$PGDAdv$ & 48 & 0.1/0.05 & 0.1/20 & -- & -- & 68.4\\
\midrule
\multirow{1}{4em}{EnsAdv}&$SGD$ & 45 & 0.1/-- & 0.1/20 & -- & -- & 71.3\\
\bottomrule
\end{tabular}}}
\vspace{3mm}
\end{table}

 \begin{table}[h]
\centering
\setlength{\tabcolsep}{1pt} 
\caption{\textbf{Hyperparameter Table for training ConvNet1/ConvNet2 models on MNIST data}}
\label{tabs5}
\centering
\resizebox{0.5\textwidth}{!}{\resizebox{\textwidth}{!}{\begin{tabular}{cccccccc}
\toprule
Model Type 
&\multicolumn{1}{p{2cm}}{\centering Training \\Method}
& Epochs
& $\eta/ \eta_{adv}$ 
& \multicolumn{1}{p{2cm}}{\centering $\eta, \eta_{adv}$ \\decay/step-size} 
& $\eta_{noise}/{\eta_{noise}}_{adv}$ 
& \multicolumn{1}{p{2cm}}{\centering $\eta_{noise}, {\eta_{noise}}_{adv}$ \\decay/step-size}
& \multicolumn{1}{p{2cm}}{\centering Test \\Accuracy in (\%)}\\
\midrule
\multirow{4}{4em}{Target ConvNet1}&$SGD$ & 100 & 0.01/-- & 0.1/50 & -- & -- & 99.1\\
&$NoL$ & 150 & 0.01/-- & 0.1/50 & 0.001/-- & 0.1/50 & 99.2\\
&$SGD_{ens}$ & 64 & 0.01/0.005 & 0.1/30 & -- & --  & 99\\
&$NoL_{ens}$ & 32 & 0.01/0.005 & 0.1/30 & 0.001/3.3e-5 & 0.1/30 & 99.1\\
&$SGD_{PGD}$ & 142 & 0.01/0.01 & 0.1/30 & -- & --  & 97.9\\
&$NoL_{PGD}$ & 162 & 0.01/0.01 & 0.1/30 & 1e-4/1e-5 & 0.1/30 & 98\\
\midrule
\multirow{2}{4em}{Source (ConvNet2)}&$SGD$ & 15 & 0.01/-- & --/-- & -- & -- & 98.6\\
&$PGDAdv$ & 128 & 0.01/0.01 & 0.1/30 & -- & -- & 97\\

\midrule
\multirow{1}{4em}{EnsAdv ConvNet1}&$SGD$ & 15 & 0.01/-- & --/-- & -- & -- & 98.8\\
\\
\bottomrule
\end{tabular}}}
\end{table}
For MNIST, we use 2 different architectures as source/ target models. ConvNet1: 32C-M-64C-M-1024FC is the model used as target. ConvNet2: 10C-M-20C-M-320FC is the model used as source.  Here, we use mini-batch SGD with momentum of 0.5, batch size 64, for training the weight parameters. Table \ref{tabs5} shows the hyperparameters used to train the models in Table 1 of main paper. The notations here are similar to that of Table \ref{tabs3}. Note, the source model trained with PGDAdv training to craft BB attacks on $NoL_{PGD}, SGD_{PGD}$ was trained with $\epsilon =0.3$,  \textit{step-size=0.01 over 40 steps}.

\subsection{Model Description for Fig. 2 in Main Paper}
\begin{table}[h]
\centering
\setlength{\tabcolsep}{1pt} 
\caption{\textbf{Hyperparameter Table for training ResNet18 models on CIFAR10  data for different types of noise modeling ($X+N, X \times N$) with all/ only negative gradient $\nabla \mathcal{L}_{N}$}}
\label{tabs6}
\centering
\resizebox{0.5\textwidth}{!}{\resizebox{\textwidth}{!}{\begin{tabular}{cccccccc}
\toprule
\multicolumn{1}{p{2cm}}{\centering Noise \\ Modeling Type} 
&\multicolumn{1}{p{2cm}}{\centering Gradient \\ $\nabla \mathcal{L}_{N}$}
& Epochs
& $\eta$ 
& \multicolumn{1}{p{2cm}}{\centering $\eta$ \\decay/step-size} 
& $\eta_{noise}$ 
& \multicolumn{1}{p{2cm}}{\centering $\eta_{noise}$ \\decay/step-size}
& \multicolumn{1}{p{2cm}}{\centering Test \\Accuracy in (\%)}\\
\midrule
$X+N$ & Negative & 120 & 0.1 & 0.1/30 & 0.001 & 0.1/30 & 78.1\\
$X+N$ & All & 120 & 0.1 & 0.1/30 & 0.001 & 0.1/30 & 77.1\\
$X\times N$ & Negative & 120 & 0.1 & 0.1/30 & 0.001 & 0.1/30 & 87.1\\
$X\times N$ & All & 120 & 0.1 & 0.1/30 & 0.001 & 0.1/30 & 85.1\\
\midrule
$SGD$ & - & 120 & 0.1 & 0.1/30 & - & - & 88.9\\
\bottomrule
\end{tabular}}}
\vspace{3mm}
\end{table}
We use mini-batch SGD with momentum of 0.9, weight decay 5e-4 and batch size 64 for training the weight parameters of the models in Table \ref{tabs6}.

\end{document}